\documentclass[conference]{IEEEtran}

\usepackage{cite}
\usepackage{amsmath,amssymb,amsfonts}
\usepackage{algorithmic}
\usepackage{graphicx}
\usepackage{textcomp}
\usepackage{xcolor}
\usepackage{url,hyperref}
\def\BibTeX{{\rm B\kern-.05em{\sc i\kern-.025em b}\kern-.08em
    T\kern-.1667em\lower.7ex\hbox{E}\kern-.125emX}}
\usepackage{subfigure}
\begin{document}

\title{Low Rank and Sparse Fourier Structure in Recurrent Networks Trained on Modular Addition
}

\author{\IEEEauthorblockN{Akshay Rangamani}
\IEEEauthorblockA{\textit{Dept. of Data Science} \\
\textit{New Jersey Institute of Technology}\\
Newark, NJ, USA \\
akshay.rangamani@njit.edu}
}

\maketitle

\begin{abstract}

Modular addition tasks serve as a useful test bed for observing empirical phenomena in deep learning, 
including the phenomenon of \emph{grokking}. Prior work has shown that one-layer transformer architectures 
learn Fourier Multiplication circuits to solve modular addition tasks. In this paper, we show that 
Recurrent Neural Networks (RNNs) trained on modular addition tasks also use a Fourier Multiplication strategy. 
We identify low rank structures in the model weights, and attribute model components to specific Fourier frequencies, 
resulting in a sparse representation in the Fourier space. We also show empirically that the RNN is robust to removing 
individual frequencies, while the performance degrades drastically as more frequencies are ablated from the model.

\end{abstract}


\section{Introduction}
Can we reverse engineer how a neural network performs the task it is trained on? How do various components of the network 
transform the inputs in order to solve the task at hand? How are the inputs represented at each layer of the network? 
\emph{Mechanistic Interpretability} \cite{cammarata2020thread:} is a growing toolkit that aims to address these questions through empirical measurements 
and causal interventions. Deep networks are often referred to as ``black-boxes'' since they consist of inscrutable piles of 
weight matrices and it is not trivial to understand what solutions we obtain. This is not unexpected, for if we knew 
\emph{a priori} what the solution should look like, we would likely not need to use machine learning to solve the problem. 
One of the goals of performing mechanistic interpretability style investigations of deep networks is to reverse engineer the 
learned solution. This may be necessary for gaining insights into problem solving strategies, to ensure safety for mission 
critical applications, and to adapt learned solutions to other settings.

Over the last five years, transformers \cite{NIPS2017_3f5ee243} have emerged as a highly effective architecture for solving 
sequence modeling problems. They have been shown to be highly effective in language modeling, in-context learning, 
few-shot learning \cite{NEURIPS2020_1457c0d6}, etc. Their computational complexity however scales quadratically with the 
size of the input sequence, in contrast with Recurrent Neural Networks (RNNs) that scale linearly. Moreover, modern recurrent 
architectures like S4 and Mamba \cite{gu2021efficiently, gu2023mamba} are also effective in similar tasks as transformers. It 
is still important to explore the similarities and differences between RNNs and transformers and understand whether they 
use similar or different strategies for solving the same task.

Understanding how deep networks learn language modeling tasks is notoriously hard due to the complex structure and the lack of complete representations for nautral language data. Instead, we choose to study deep networks trained on simple algorithmic tasks like modular addition, where we have a good understanding of what a good solution should look like.
Recently, Nanda et al. performed a mechanistic interpretability analysis of a small one layer transformer trained on a modular addition 
task and uncovered a Fourier multiplication algorithm encoded in the weights and activations of the model \cite{nanda2023progress}. 
They also used this modular addition circuit to track the progress of the model through a phase called \emph{grokking} where the model 
has fit the training data, but does not generalize to unseen examples. In this abstract, we train an RNN on the same modular addition 
problem and show that a Fourier multiplication algorithm can be uncovered from their its and activations as well. We also make causal 
interventions into the weights of the model and show that these Fourier frequencies are causally related to achieving high accuracy on 
the modular addition problem.

\subsection{Related Work}
In the modern deep learning literature interest in learning simple algorithmic tasks like modular addition was sparked by 
Power et al. \cite{power2022grokking}, in which the authors identified a phenomenon called \emph{grokking} where models go through 
a phase of memorizing a task on the training dataset while not generalizing to test data. Since then, researchers have shown 
interest in explaining the phenomenon of grokking in follow up papers \cite{varma2023explaining, morwani2024feature, mohamadi2024you, mallinar2024emergence}. 
There has also been follow up work in characterizing the solutions obtained on modular addition tasks by feedforward networks 
like multilayer perceptrons with quadratic activations \cite{doshi2024grokking}, and transformers \cite{nanda2023progress, zhong2024clock}. 
Some theoretical results outlining the existence of Fourier frequencies in the solutions to modular addition tasks have 
also been recently developed \cite{marchetti2024harmonics, morwani2024feature}. Our paper focuses on characterizing the final 
learned representations in recurrent networks, as opposed to transformers or feedforward networks as described above. 
We also highlight the sparsity of the Fourier features, the low rank nature of the solutions obtained, and the relationship betwen 
the two dimensions of sparse structure in deep networks.

\section{Experimental Setup}
\begin{figure}
    \centering
    \includegraphics[scale=0.5]{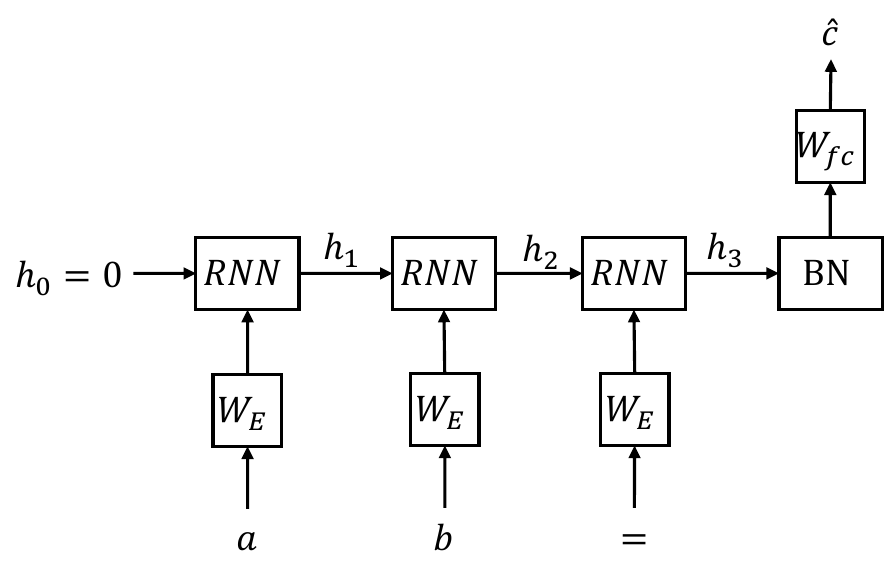}
    \caption{Unrolled computational graph of the RNN trained to perform modular addition}
    \label{fig:diagram}
\end{figure}

We train RNNs to solve the addition modulo $113$ task. We feed input sequences of the form $|a|b|=|$ and train the RNN 
to predict the token $c=a+b (\textrm{mod } p)$ (where $p=113$). We generate all $113 \times 113 = 12769$ sequences and choose a random $30\%$ 
fraction of these to train the model. Our model is a single layer vanilla RNN with a state of size $d_h=256$ using the 
tanh nonlinearity. The input tokens are passed through a trainable embedding layer ($W_E$) before the RNN. The last hidden 
state of the RNN is normalized, and finally passed through an unembedding ($W_{fc}$) layer to determine the output token $\hat{c}$. 
The embedding and unembedding layers are not tied to each other. We use the Adam optimizer with full batches and a learning 
rate of $0.01$ and weight decay of $5\times 10^{-5}$ to train the model. We use full batch descent, i.e., the gradient is 
computed using the entire training set at each iteration. We train for $20,000$ epochs and observe that the model achieves 
zero test loss after around $4000$ epochs. The test loss and accuracy is evaluated on the remaining $70\%$ fraction of the dataset.

\begin{figure*}[htbp!]
     \centering
     \begin{subfigure}
         \centering
         \includegraphics[scale=0.5]{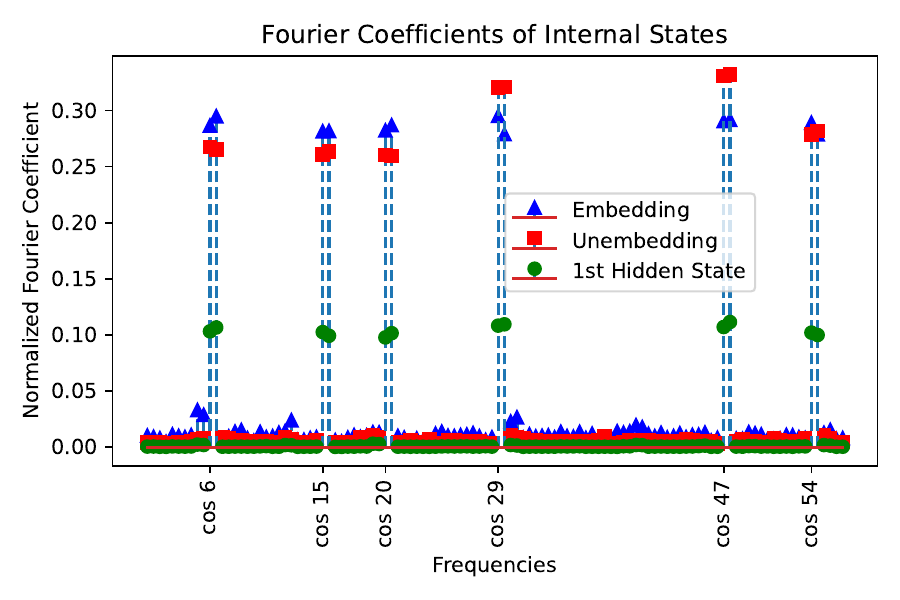}
     \end{subfigure}
     \hfill
     \begin{subfigure}
         \centering
         \includegraphics[scale=0.5]{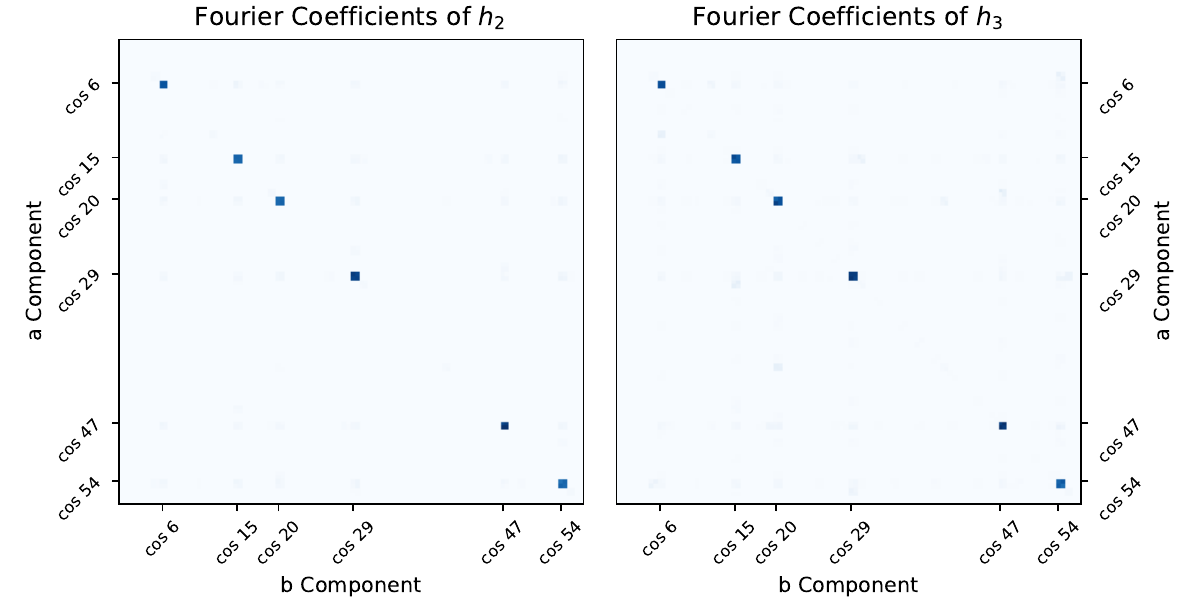}         
     \end{subfigure}
     \caption{Tracking the Fourier coefficients of every node in the computational graph. On the left we plot the coefficients of the embedding and unembedding layers, as well as $h_1$ since it is just a function of $a$. On the right we plot the coefficients of the hidden states $h_2, h_3$ which are functions of $a,b$ (darker colors indicate larger magnitude coefficients). For all tensors, we find that the same $6$ frequencies, $\omega_k = \frac{2\pi k}{P}$ for $k \in \{6, 15, 20, 29, 47, 54\}$ are the only significant coefficients.}
     \label{fig:fourier_coeffs}
\end{figure*}

\section{Results}
We obtain a trained network using the procedure outlined in the previous section. We now probe this model using 
mechanistic interpretability techniques. Taking inspiration from \cite{nanda2023progress}, we perform a 
Fourier domain analysis of every node in the computational graph shown in figure \ref{fig:diagram} and find a 
sparse Fourier representation at every node. We then study the singular value spectra of the model weights, and 
uncover a low rank structure. We are also able to find the Fourier coefficients of singular vectors of the unembedding 
layer and connect each singular vector to a specific frequency. Finally, we perform ablations of the different 
Fourier frequencies and observe the change in model accuracy.

\subsection{Fourier Spectra of Inputs, Hidden States, Outputs}

In this subsection we study Fourier coefficients of different nodes in computational graph of figure \ref{fig:diagram}.
Similar to the procedure in \cite{nanda2023progress} we construct a matrix $F \in \mathbb{R}^{p \times p}$ 
consisting of a constant row, and $ \left\{ \textrm{cos}\left( \frac{2 \pi kn}{p}\right) \right\}_{n=1}^{p}, \left\{ \textrm{sin}\left( \frac{2 \pi kn}{p}\right) \right\}_{n=1}^{p}$ for $k=1,\ldots,56$ ($k \leq \lfloor \frac{p-1}{2} \rfloor$ are the unique Fourier frequencies; larger values of $k$ simply capture harmonics of these frequencies). 
We multiply the embedding matrix $W_E \in \mathbb{R}^{p \times d_h}$ and the unembedding matrix 
$W_{fc} \in \mathbb{R}^{d_h \times p}$ by $F$ along the appropriate dimensions to obtain their Fourier coefficients. 
Since the first hidden state $h_1$ is only a function of one input, we collect $h_1$ evaluated over all 
inputs $a$ into a $\mathbb{R}^{p \times d_h}$ matrix. The hidden states $h_2,h_3$ are functions of both 
inputs $a,b$ and hence can be decomposed in Fourier bases along both input axes. The same matrix $F$ is 
applied along both axes to obtain the 2D Fourier coefficients. The results of these computations are presented 
in figure \ref{fig:fourier_coeffs}.

In the left panel of figure \ref{fig:fourier_coeffs}, we plot the Fourier coefficients (normalized to unit $\ell_2$ norm) 
of the embedding layer, unembedding layer, and $h_1$. The 2D Fourier coefficients of $h_2, h_3$ are plotted in the subsequent two panels. Most of the energy in the signals is evidently contained in $6$ frequencies $\omega_k = \frac{2\pi k}{p}$ for $k \in \{6, 15, 20, 29, 47, 54\}$ as shown in Figure \ref{fig:fourier_coeffs}. Moreover we note that the dominant frequencies are exactly the same for all signals. This correspondence emerges even though we do not tie the weights $W_E, W_{fc}$ when training. We also observe that the hidden states $h_2, h_3$ of the RNN after seeing both the $a,b$ tokens are also sparse in the Fourier domain. These frequencies correspond to the exact frequencies used by the other layers. 

\subsection{Low Rank Structure in Model Weights and connections to Fourier Spectra}

\begin{figure}
    \centering
    \includegraphics[scale=0.5]{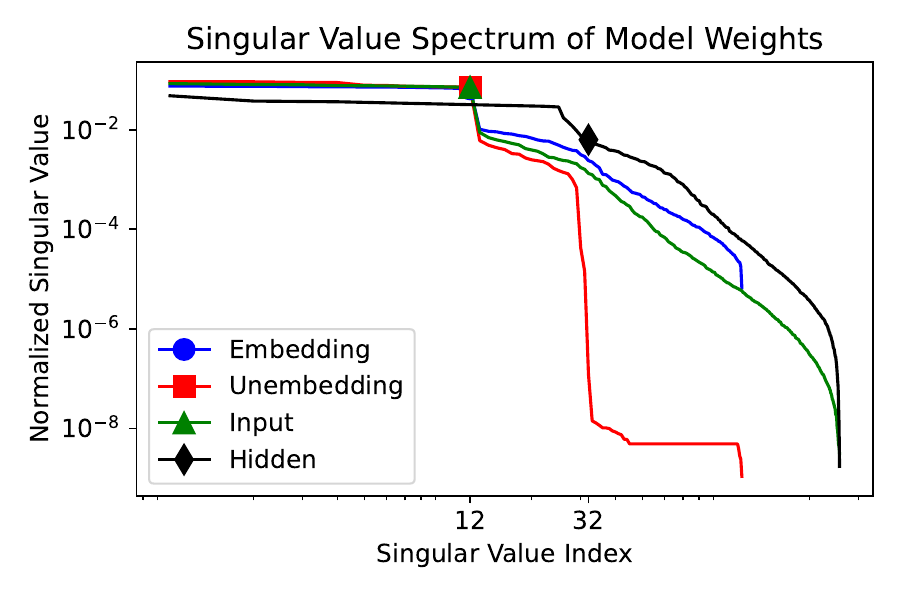}
    \caption{Singular value spectra of model weights are low rank. The number of components $r$ required to maintain $100\%$ model accuracy as follows: embedding matrix $W_E$ : $r=12$ containing $85.4\%$ of the energy , unembedding matrix $W_{fc}$ : $r=12$ containing $95.5\%$ of the energy, input-hidden matrix $W_{ih}$ : $r=12$ containing $91.4\%$ of the energy, and hidden-hidden matrix $W_{hh}$ : $r=32$ containing $89.4\%$ of the energy.}
    \label{fig:rank}
\end{figure}

In the previous subsection we studied the Fourier spectra of the different signals computed in the model. Now we turn our attention to the structure of the weights. We compute the singular value decompositions (SVDs) of each matrix and plot the singular value spectra in figure \ref{fig:rank}. By varying the number of components included in each weight matrix and measuring the model accuracy, we find that the embedding ($W_E$), unembedding ($W_{fc}$), and input-hidden ($W_{ih}$) weight matrices have only $r=12$ significant components while the hidden-hidden ($W_{hh}$) weight matrix has $r=32$ significant components. If we compute the energy contained in these components (measured as $\sum_{i=1}^r \sigma_i / \sum_{j=1}^{n} \sigma_j$, with $n=p, d_h$ depending on the context, the amount varies for the different matrices. We find it is $85.4\%$ for $W_E$, $95.5\%$ for the $W_{fc}$, $91.4\%$ for $W_{ih}$, and $89.4\%$ for $W_{hh}$. However, these are the minimal number of components required to achieve $100\%$ performance on the modular addition task. This low rank structure indicates that the model is not using the full $d_h = 256$-dimensional space to solve the task. In contrast if we restrict the model weights to the orthogonal complement of the significant subspaces - i.e., only include the components $r>12/32$ for the respective model weights, the accuracy of the model drops to the chance accuracy of $0.885\%$.

\subsection{Aligning Singular Vectors with Fourier Frequencies}

\begin{figure*}
    \centering
    \includegraphics[scale=0.5]{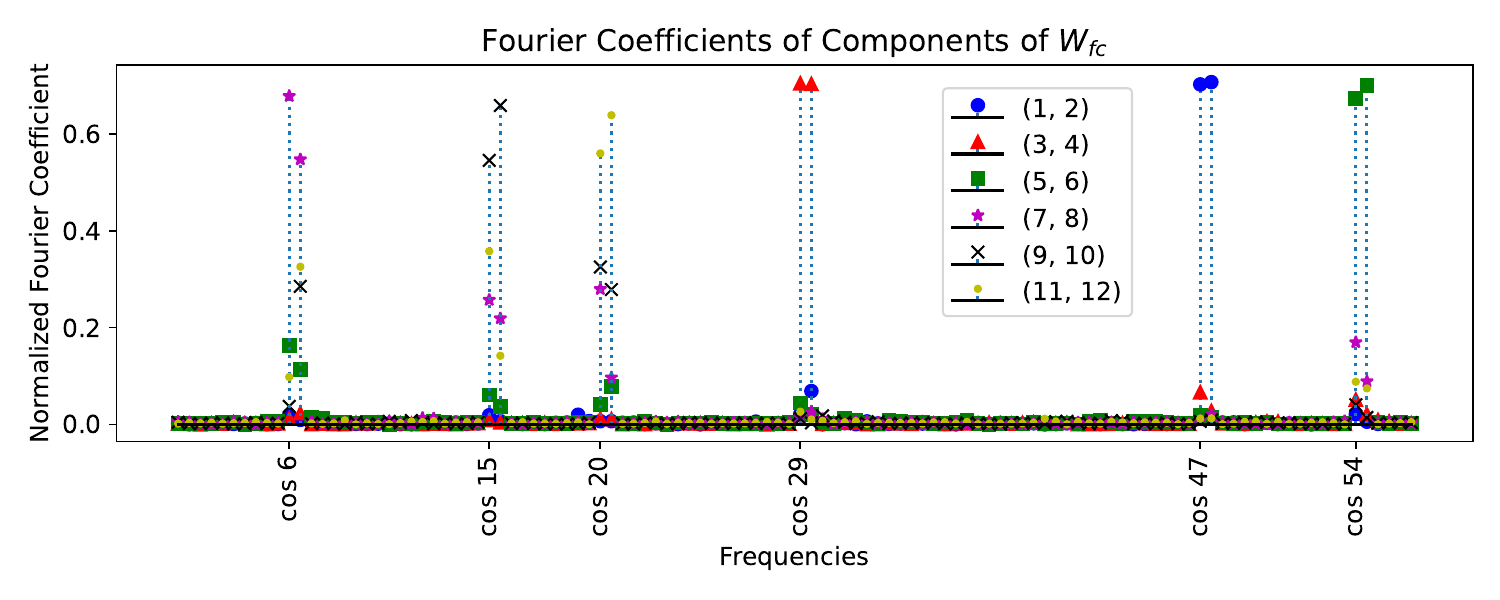}
    \caption{Identifying the Fourier frequencies associated with different singular vectors $U_{fc}^k, V_{fc}^k$ for $k=1,\ldots,12$ of the unembedding matrix. We find that each consecutive pair of singular vectors is associated with a distinct frequency.}
    \label{fig:fc_components}
\end{figure*}

\begin{table}
\caption{Model Accuracy after Frequency Ablation}
\begin{center}
\begin{tabular}{|c|c|c|}
\hline
Ablated Frequency & Single Frequency $\omega_k$ & Multiple Frequencies $\omega_{k_1:k_n}$ \\ \hline
$k_1=47$ &  $100\%$ & $100\%$ \\ \hline
$k_2=29$ & $99.9\%$ & $100\%$ \\ \hline
$k_3=54$ & $100\%$ & $87.3\%$ \\ \hline
$k_4=6$ & $99.4\%$ & $28.3\%$ \\ \hline
$k_5=15$ & $100\%$ & $6.2\%$ \\ \hline
$k_6=20$ & $99.8\%$ & $0.885\%$ \\ \hline
\end{tabular}
\label{tab:ablation}
\end{center}
\end{table}

In this subsection we perform a fine-grained analysis of the singular vector components of the unembedding matrix $W_{fc}$. We restrict $W_{fc}$ to pairs of singular vectors $U_{fc}, V_{fc}$ and compute the Fourier coefficients of the resulting low rank approximation. We find that the $12$ significant singular vectors can be grouped into $6$ pairs. The Fourier coefficients of each pair are dominated by a single frequency $\omega_k = \frac{2\pi k}{p}$ for $k \in \{6, 15, 20, 29, 47, 54\}$. Thus we can confirm that the rank of $W_{fc}$, $r=12$ and the number of significant Fourier coefficients $6 \times 2$ are deeply related, with different singular vectors corresponding to different frequencies. This may also be the reason for $W_E$ and $W_{ih}$ concentrating their energy in a low-dimensional $r=12$ subspace.

Having identified the frequencies associated with singular vector components, we can perform another ablation test. We can remove different model frequencies, and observe changes in the model performance. We present these results in table \ref{tab:ablation}. In the first column of the table we report the results of removing just one of the frequencies $\omega_k =\frac{2\pi k}{p} $. We see that individual frequencies are not singularly important. Even if one frequency is removed, the model is able to compute the correct answer using the other frequencies. However, removing multiple frequencies from the model does drastically reduce the performance. We report the results of removing $k_1, \ldots, k_6$ frequencies and observe that the model performance drops monotonically to chance accuracy with the removal of subsequent model frequencies.

We can extend this line of inquiry into the other layers of the model, and find the Fourier domain representation of $W_E, W_{ih}, W_{hh}$. However in these layers we do not find a one-one mapping between singular vectors and frequencies. For instance, we find that $\mathcal{F}\{U_E^1 V_E^{1\top} \} \approx 0.65 \textrm{sin} \left(\frac{2\pi \times 47n}{p} \right) + 0.4 \textrm{cos} \left(\frac{2\pi \times 47n}{p} \right) + 0.5 \textrm{cos} \left(\frac{2\pi \times 29n}{p} \right)$. Similarly other singular vector components of the respective weight matrices have a few dominant frequencies, but we do not find a clear relationship between particular singular vectors and particular frequencies. While ablating all components related to a particular frequency does remove that frequency from the Fourier spectra of all nodes in the model's computational graph, it also diminishes other components, and affects the overall model accuracy. The fine-grained control of model frequencies is most effective at the unembedding layer $W_{fc}$, even though all layers have sparse Fourier spectra.

\subsection{Exact Implementation of Fourier Multiplication}
\begin{table}
\caption{Fitting Fourier representations of $V^{k\top}_{fc}h_3$ to $\alpha_k \textrm{cos} \omega_k (a+b) + \beta_k \textrm{sin} \omega_k (a+b)$}
\begin{center}
\begin{tabular}{|c|c|}
\hline
Frequency & Relative Error of Fit \\ \hline
$k_1=47$ &  $7.42 \times 10^{-3}$ \\ \hline
$k_2=29$ & $4.65 \times 10^{-3}$ \\ \hline
$k_3=54$ & $6.82 \times 10^{-3}$ \\ \hline
$k_4=6$ & $3.32 \times 10^{-3}$ \\ \hline
$k_5=15$ & $1.69 \times 10^{-2}$ \\ \hline
$k_6=20$ & $1.39 \times 10^{-2}$ \\ \hline
\end{tabular}
\label{tab:rel_error}
\end{center}
\end{table}

In the previous subsections we saw that all weights and hidden states of the RNN are sparse in the Fourier domain. However in order to conclusively establish that our model uses the Fourier multiplication algorithm, we need to confirm that it computes $\textrm{cos} \omega_k (a+b)$ and $\textrm{sin} \omega_k (a+b)$ in the last layer. We normalize the final hidden state $h_3$ and project it onto the singular vectors $V^k_{fc}$ of $W_{fc}$ that correspond to different frequencies $\omega_k$. We find the best fit of these Fourier representations against $\alpha_k \textrm{cos} \left(\omega_k (a+b) \right) + \beta_k \textrm{sin} \left( \omega_k (a+b) \right)$ and report the relative errors for these fits in table \ref{tab:rel_error}. We find that all relative errors are all $\lesssim 10^{-2}$. This establishes that the algorithm used by the RNN to solve the modular addition task is to project each input token $a$ into an embedding space $[\textrm{cos}( \omega_k a), \textrm{sin}( \omega_k a)]$ for a sparse set of frequencies $\omega_k$. The input-hidden and hidden-hidden weight matrices embed the signal $[\textrm{cos}\left( \omega_k (a+b)\right), \textrm{sin}\left( \omega_k (a+b)\right)]$ in the hidden state, which can be decoded by the unembedding matrix to produce the correct result. By probing every node of the computational graph of the RNN, we have obtained a complete description of how an RNN solves a modular addition task.

\subsection{Other Training Runs}

While we focus on a single model to illustrate our investigation, we were also able to replicate our findings across different training runs, and training with more data. We varied that fraction of the total dataset used from $30\%$ to $70\%$ of the entire dataset in increments of $10\%$. In all cases we found the same low rank structure and sparse Fourier spectra. The number of significant Fourier frequencies varied between $6$ and $8$, and the corresponding rank of $W_{fc}$ and $W_E, W_{ih}$ was always $2\times$ the number of significant frequencies. The interpretability of singular vector components of $W_{fc}$ also varied, though there were always a few dominant frequencies associated with each singular vector component.

\section{Discussion, Conclusion, and Future Work}

In the previous section we have shown that the Fourier multiplication algorithm for modular addition appears in RNNs. This algorithm has also been described in one-layer transformers \cite{nanda2023progress} and multilayer perceptrons with quadratic activations \cite{gromov2023grokking}. Prior work however does not consider RNNs, and does not provide the sparse, low rank structure described here especially in the case of MLPs. Most theoretical results \cite{marchetti2024harmonics, morwani2024feature, mohamadi2024you} do not identify this structure. They only construct a network that uses all Fourier frequencies, and show that such a network can solve the modular addition task. Our paper on the other hand shows that RNNs, in addition to one-layer transformers, learn a solution that is low rank, and sparse in Fourier space. Moreover we identify the components of the weight matrices that are related to specific frequencies. We believe this description in RNNs has not been previously identified, and identifying the factors in training that result in a sparse, low rank structure is an intriguing task for future investigation. While learning modular arithmetic is in some sense a toy problem, understanding deep learning in this setting can lead to insights for learning more interesting group operations and transformations that arise in more practical settings. 

\section*{Acknowledgments}
We thank Lakshya Chauhan for running experiments for an early version of this paper.

\pagebreak
\bibliographystyle{IEEEtran}
\bibliography{references}

\end{document}